\documentclass[11pt,english]{article}

\usepackage{latexsym}
\usepackage{epsfig}
\usepackage{amssymb}
\usepackage{array}
\usepackage{float}
\usepackage{algorithm}
\usepackage{algorithmic}
\graphicspath{{figures/}{./}}

\newenvironment{proof}{{\bf Proof. } }{{\hfill $\Box$}\vspace{.5pc}}
\newtheorem{theorem}{Theorem}

\newtheorem{definition}[theorem]{Definition}
\newtheorem{lemma}[theorem]{Lemma}

\newtheorem{remark}[theorem]{Remark}

\newtheorem{conjecture}[theorem]{Conjecture}
\floatstyle{ruled}

\floatstyle{ruled}
\newfloat{algo}{thp}{lop}
\floatname{algo}{Algorithm}

\newcommand{\IF}[1]{\textbf{if} {#1}}
\newcommand{\THEN}{\textbf{then} }
\newcommand{\ELSE}{\textbf{else} }

\newcommand{\AB}{<\!A\! \leadsto\! B\!>} 
\newcommand{\BQ}{<\!B\! \leadsto\! Q\!>} 
\newcommand{\QaQ}{<\!Q\! \leadsto\! {\tt a}Q\!>} 
\newcommand{\aQN}{<\!{\tt a}Q\! \leadsto\!N{\tt gon}\!>} 
\newcommand{\Ngon}{<\!A\! \leadsto\!N{\tt gon}\!>} 

\newcommand{\BEGLIST}{\begin{list}{}{\partopsep -3pt \parsep -2pt \listparindent -0pt \labelwidth .5in}}
\newcommand{\ENDLIST}{\end{list}}

\begin{document}
\title{Circle Formation of Weak Mobile Robots
\thanks{Independantly of our work, in~\cite{PPS06}, the authors show the validity of the Conjecture of 
D\'efago \& Konagaya. This part has been removed from the submitted version of this technical report.}}

\author{Yoann Dieudonn\'e$^1$ \and Ouiddad Labbani-Igbida$^2$ \and Franck Petit$^1$ \\
\begin{tabular}[t]{c@{\extracolsep{8em}}c}
$^1$ LaRIA, CNRS FRE 2733 & $^2$ CREA \\
Universit\'{e} de Picardie Jules Verne & Universit\'{e} de Picardie Jules Verne\\
France & France
\end{tabular}}
\date{}
\maketitle

\begin{abstract}
The contribution is twofold.  We first show the validity of the conjecture of D\'efago and Konagaya 
in \cite{DK02}, i.e., there exists no deterministic oblivious algorithm solving the Uniform Transformation
Problem for any number of robots$^*$. 
Next, a protocol which solves deterministically the Circle Formation Problem in finite time 
for any number $n$ of weak robots---$n \notin \{4,6,8\}$---is proposed.  
The robots are assumed to be uniform, anonymous, 
oblivious, and they share no kind of coordinate system nor common sense of direction. 
\medskip

\end{abstract}

\section{Introduction}

In this paper, we address the class of distributed systems where computing units are
\emph{autonomous} \emph{mobile robots} (also sometimes referred to \emph{sensors} or 
\emph{agents}), i.e., devices equipped with sensors which do not depend on a central 
scheduler and designed to move in a two-dimensional plane. 
Also, we assume that the robots cannot remember any previous observation nor computation performed 
in any previous step.  Such robots are said to be \emph{oblivious} (or \emph{memoryless}). 
The robots are also \emph{uniform} and \emph{anonymous}, i.e, they 
all have the same program using no local parameter (such that an identity) 
allowing to differentiate any of them.
Moreover, none of them share any kind of common coordinate mechanism or common 
sense of direction, and they communicate only by observing the position of the others.

The motivation behind such a weak and unrealistic model is the study of the minimal level of ability 
the robots are required to have in the accomplishment of some basic cooperative tasks in a deterministic 
way, e.g.,~\cite{SS90,SY99,FPSW99,P02}.  
Among them, the \emph{Circle Formation Problem} (CFP) has received a particular attention. 
The CFP consists in the design of a protocol insuring that starting from an initial arbitrary configuration, all $n$ robots 
eventually form a circle with equal spacing between any two adjacent robots.  In other words, the robots are required to 
form a \emph{regular $n$-gon} when the protocol terminated.  

\paragraph{Related Works.}

An informal CFP algorithm is presented in~\cite{D95} to show 
the relationship between the class of pattern formation algorithms and the concept of self-stabilization
in distributed systems~\cite{D00}.
In~\cite{SS96}, an algorithm based on heuristics is proposed for the formation of a circle approximation.
A CFP protocol is given in~\cite{SY99} for non-oblivious robots with an unbounded memory.  
Two deterministic algorithms are provided in~\cite{DK02,CMN04}.  In the former work, the robots
asymptotically converge toward a configuration in which they are uniformly distributed on the boundary 
of a circle.  This solution is based on an elegant Voronoi Diagram construction.  
The latter work avoid this construction by making an extra assumption on the initial position of robots. 
In~\cite{DP06}, properties on Lyndon words are used to achieve a Circle Formation Protocol (the exact
$n$-gon is eventually built) for a prime number of robots.  
All the above solutions work in the semi-asynchronous model introduced in~\cite{SY96}.
The solution in~\cite{K05} works in a fully asynchronous model, but when $n$ is even, 
the robots may only achieve a biangular circle---the distance between two adjacent robots is 
alternatively either $\alpha$ or $\beta$.  

A common strategy in order to solve a non trivial problem as CFP is to combine subproblems which are easier 
to solve.  In general, CFP is separated into two distinct parts:
The first subproblem consists in placing the robots along the boundary of a circle $C$, without considering their relative 
positions. The second subproblem, called \emph{uniform  transformation problem} (UTP), consists in starting from 
there, and arranging robots, without them leaving the circle $C$, evenly along the boundary of $C$. 
In~\cite{DK02}, the authors present an algorithm, for the second subproblem which converges toward a 
homogeneous distribution of robots, but it does not terminate deterministically. By the way, they conjecture 
that there is no deterministic solution solving UTP in finite time in the semi-asynchronous model
in~\cite{SY96}---the robots being uniform, anonymous, oblivious, and none of them 
sharing any kind of coordinate system or common sense of direction.  

\paragraph{Contribution.}

The contribution is twofold.  We first show the validity of the conjecture of D\'efago and Konagaya in \cite{DK02},
i.e., there exists no deterministic oblivious algorithm solving the Uniform Transformation
Problem for any number of robots\footnote{Independantly of our work, in~\cite{PPS06}, the authors 
show the validity of the Conjecture of D\'efago \& Konagaya. 
This part has been removed from the submitted version of this technical report.}.
Next, we propose the first protocol which solves deterministically CFP in finite time 
for any number $n$ of weak robots, 
provided that $n \notin \{4,6,8\}$.  By weak, we mean that the robots are assumed to be uniform, anonymous, 
oblivious, and they share no kind of coordinate system nor common sense of direction. 
Our protocol is not based on UTP, but it is based on concentric circles formed by the robots. 

\paragraph{Outline of the Paper.}	 

In the next section (Section~\ref{sec:model}), we describe the distributed 
systems and the model we consider in this paper.  In the same section, we present 
the problem considered in this paper.  
Section~\ref{sec:conjDK} addresses the conjecture of D\'efago and Konagaya. 
The algorithm is proposed in Section~\ref{sec:algo}.
Finally, we conclude this paper in Section~\ref{sec:conclu}.

\section{Preliminaries}
\label{sec:model}

In this section, we define the distributed system, basic definitions and the problem considered in this paper.  

\paragraph{Distributed Model.}

We adopt the model introduced~\cite{SY96}, in the remainder referred as~$SSM$.  
The \emph{distributed system} considered in this paper consists of $n$ robots  
$r_{1}, r_{2},\cdots , r_{n}$---the subscripts $1,\ldots ,n$ are used for notational purpose only.
Each robot $r_{i}$, viewed as a point in the Euclidean plane, move on this two-dimensional 
space unbounded and devoid of any landmark.  When no ambiguity arises, $r_{i}$ also denotes the 
point in the plane occupied by that robot. 
It is assumed that the robots never collide and that two or more robots may 
simultaneously occupy the same physical location. 
Any robot can observe, compute and move with infinite decimal precision.
The robots are equipped with sensors allowing to detect the instantaneous position of the other robots in the plane.  
Each robot has its own local coordinate system and unit measure.  
The robots do not agree on the orientation of the axes of their local coordinate system, 
nor on the unit measure. 
They are \emph{uniform} and \emph{anonymous}, i.e, they all have the same program using no 
local parameter (such that an identity) 
allowing to differentiate any of them.  
They communicate only by observing the position of the others and they are \emph{oblivious}, i.e.,
none of them can remember any previous observation nor computation performed 
in any previous step. \\

Time is represented as an infinite sequence of time instant $t_0, t_1, \ldots, t_j, \ldots$ 
Let $P(t_j)$ be the multiset of the positions in the plane occupied by the $n$ 
robots at time $t_j$ ($j\geq0$). For every $t_j$, $P(t_j)$ is called the \emph{configuration} 
of the distributed system in $t_j$.
$P(t_j)$ expressed in the local coordinate system of any robot $r_i$ is called a \emph{view}, denoted 
$v_i(t_j)$.
At each time instant $t_j$ ($j\geq 0$), each robot $r_i$ is either {\it active} or {\it inactive}. 
The former means that, during the computation step $(t_j,t_{j+1})$, using 
a given algorithm, $r_i$ computes in its local coordinate system a position $p_i(t_{j+1})$ depending 
only on the system configuration at $t_j$, and moves towards $p_i(t_{j+1})$---$p_i(t_{j+1})$ can be equal to
$p_i(t_j)$, making the location of $r_i$ unchanged.
In the latter case, $r_i$ does not perform any local computation and remains at the same position. 

The concurrent activation of robots is modeled by 
the interleaving model in which the robot activations are driven by a \emph{fair scheduler}.  
At each instant $t_j$ ($j\geq 0$), the scheduler 
arbitrarily activates a (non empty) set of robots.  
Fairness means that every robot is infinitely often activated by the scheduler.

\paragraph{The Circle Formation Problem.}

In this paper, the term ``\emph{circle}'' refers a circle having a radius strictly greater than zero.  
Consider a configuration at time $t_k$ ($k\geq 0$) in which the positions of the $n$ 
robots are located at distinct positions on the circumference of a 
circle $C$. 
At time $t_k$, the \emph{successor} $r_j$, $j\in 1\ldots n$, 
of any robot $r_i$, $i \in 1 \ldots n$ and $i \neq j$, is 
the single robot such that no robot exists between $r_i$ and $r_j$ 
on $C$ in the clockwise direction.
Given a robot $r_i$ and its successor $r_j$ on $C$ centered in $O$:\newline
\begin{enumerate} 
\item $r_i$ is said to be the \emph{predecessor} of $r_j$;
\item $r_i$ and $r_j$ are said to be \emph{adjacent};
\item $\widehat{r_i O r_j}$ denotes the angle centered in $O$ and 
with sides the half-lines $[O,r_i)$ and $[O,r_j)$ such that no
robots (other than $r_i$ and $r_j$) is on $C$ inside $\widehat{r_i O r_j}$.
\end{enumerate}

\begin{definition}[regular $n$-gon] A cohort of $n$ robots ($n \geq 2$) forms 
(or is arranged in) a regular $n$-gon if the robots 
take place on the circumference of a circle $C$ centered in $O$
such that for every pair $r_i,r_j$ of robots, if $r_j$ is the successor of $r_i$ on $C$,
then $\widehat{r_i O r_j} = \delta$, where $\delta= \frac{2\pi}{n}$.  
The angle~$\delta$ is called the \emph{characteristic angle} of the $n$-gon. 
\end{definition}

The problem considered in this paper, called CFP (\emph{Circle Formation Problem}) 
consists in the design of a 
distributed protocol which arranges a group of $n$ ($n > 2$) 
mobile robots with initial distinct positions  
into a \emph{regular $n$-gon} in finite time.
(We ignore the trivial cases $n\leq 2$ because in that cases, they always form a regular $n$-gon.)

\section{On The Conjecture of D\'efago and Konagaya}
\label{sec:conjDK}

\begin{definition}[UTP Algorithm]
A distributed algorithm $A$ solves the uniform transformation problem (UTP) if and only if, 
starting from a configuration where the robots are arbitrarily located along the 
circumference of a circle $C$, $(i)$ none of the robots leaves the circumference 
of $C$ during the execution of $A$ and, $(ii)$ all the robots eventually form a regular $n$-gon.
\end{definition}

In this section, we show the validity of the following conjecture:

\begin{conjecture}[\cite{DK02}]
\label{conj:DK02}
There exists no deterministic oblivious algorithm solving UTP in SYm for any number of robots.
\end{conjecture}

\begin{figure}[!htbp]
\begin{center}
    \epsfig{file=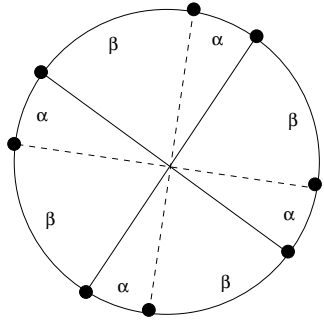, width=0.25\linewidth}
 \caption{An example showing a strict biangular circle ($\alpha \neq \beta$). \label{fig:biangu}}
\end{center}
\end{figure}

\begin{definition}[Biangular circle] A cohort of $n$ robots ($n \geq 2$) forms 
(or is arranged in) a biangular circle if the robots  
take place on the circumference of a circle $C$ centered in $O$
and there exist two non zero angles $\alpha, \beta$ such that 
for every pair $r_i,r_j$ of robots, if $r_j$ is the successor of $r_i$ on $C$,
then $\widehat{r_i O r_j} \in \{\alpha,\beta\}$ and $\alpha$ and $\beta$ alternate
in the clockwise direction.
\end{definition}
\begin{remark}
\label{rem:biangular}
In a biangular circle, $\alpha+\beta=4\frac{\pi}{n}$.
\end{remark}
Obviously, if $\alpha=\beta$ then, the $n$ robots form a regular $n$-gon, and $n$ can
either odd or even.  If $\alpha \neq \beta$, then $n$ must be even ($n=2p,\ p \geq 1$).  
In that case, the biangular circle is called a \emph{strict} biangular circle---refer to
Figure~\ref{fig:biangu}.  In that case, there exist two distinct groups $G_1$ and $G_2$ such that: 
\begin{enumerate}
 \item $|G1|=|G2|=\frac{n}{2}$;
 \item The $\frac{n}{2}$ robots in $G_1$ (resp. $G_2$) form a regular $\frac{n}{2}$-gon;
 \item The robots do not form a regular $n$-gon. 
\end{enumerate}

Given a configuration $P(t_j)$, if the $n$ robots form a strict biangular circle, then 
$G_1(t_j)$ (resp. $G_2(t_j)$) indicates the positions of the robots in $G_1$ (resp. $G_2$) at 
time $t_j$.

The idea of proof is as follows: we show, for each any strategy of 
the robots, there exists a particular activation schedule foiling it. More precisely, 
we show that if initially the robots form a strict biangular circle, then they may not eventually form a 
regular $n$-gon in a deterministic way because of the unpredictability of the activation schedule. 
This result holds for the case $n=2p$ ($p > 1$).  So, it proves the general result.

\begin{lemma}
\label{lem1:conj}
Let $A$ be a deterministic oblivious algorithm solving $UTP$ in finite time and  
a configuration $P(t_j)$ such that the $n$ robots form a strict biangular circle in $P(t_j)$.  
If any robot $r_i$ becomes active at time $t_j$, by executing $A$, it moves toward a 
position $p(t_{j+1}$ such that $p_i(t_j) \neq p_i(t_{j+1})$.
\end{lemma}

\begin{figure}[!htbp]
\begin{center}
    \epsfig{file=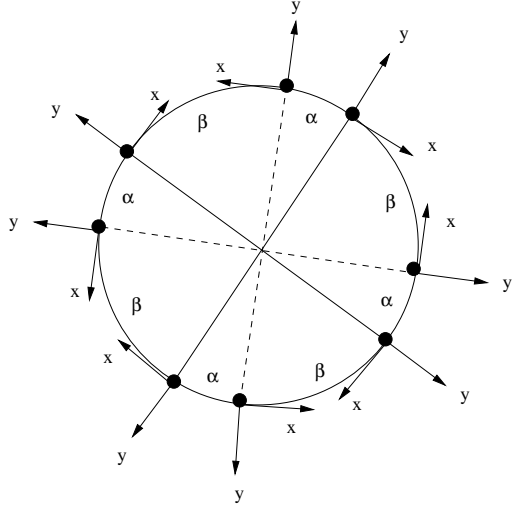, width=0.4\linewidth}
 \caption{An example showing the initial configuration in the proof of Lemma~\ref{lem1:conj}. \label{fig:biangu2}}
\end{center}
\end{figure}

\begin{proof}
Since the robots have no common coordinate system and sense of direction, then all of them 
may have the same view at $t_j$, i.e., $v_i(t_j)=v_k(t_j)$, for all $r_i$,$r_k$.  Such a configuration
is shown in Figure~\ref{fig:biangu2}.
Assume by contradiction that, there exists a robot $r_i$ which becomes active at 
$t_j$ and move toward a position $p(t_{j+1})$ such that $p_i(t_j) \neq p_i(t_{j+1})$.
Since $A$ is a deterministic algorithm, if all the robots are the same view, then
all the active robots choose the same behavior, i.e., $\forall r_i$ such that $r_i$ is 
active at $t_j$, $r_i$ move to a position $p(t_{j+1})$ such that $p_i(t_j) \neq p_i(t_{j+1})$. 
 From the model, all the inactive robots remain at the same position at time $t_{j+1}$.  So, 
$P(t_j)=P(t_{j+1}$.  Since the robots are oblivious and $A$ is a deterministic algorithm, we can easily 
deduce by induction (starting from $t_j$) that the robots always form a stric biangular circle by 
executing $A$. 
\end{proof}

\begin{lemma}
\label{lem2:conj}
Let $A$ be a deterministic oblivious algorithm solving $UTP$ in finite time and  
a configuration $P(t_j)$ such that the $n$ robots form a strict biangular circle in $P(t_j)$.
If any robot $r_i$ becomes active at time $t_j$, by executing $A$, it moves toward a 
position $p_i(t_{j+1})$ such that $p_i(t_{j+1}) \ne p_k(t_j)$ for all $r_k\ne r_i$.
\end{lemma}

\begin{proof}
By contradiction, assume that there exists $r_i,\ r_k$ such that $r_i$ moves toward a 
position $p_i(t_{j+1})$ such that $p_i(t_{j+1}) = p_k(t_j)$. 
Clearly, if $r_k$ is inactive at time $t_j$, $r_i$ and $r_k$ have the same position at time $t_{j+1}$. 
Assume that $r_i$ and $r_k$ have the same coordinate system. So, they share the same view, i.e., 
$v_i(t_{j+1})=v_k(t_{j+1})$.  Assume that from $t_{j+1}$ on, $r_i$ and $r_k$ are always active at the same time. 
So, from $t_{j+1}$ on, for any move that $r_i$ makes by executing $A$, $r_k$ makes the same move as $r_i$. 
Therefore, at time $t_{j+2}$, $r_i$ and $r_k$ are again located at the same point and they share the same view. 
By induction, starting from $t_{j+1}$, it could be impossible to separate $r_i$ and $r_k$ in a deterministic manner. 
Hence, the $n$-gon cannot be eventually formed and $A$ is not a deterministic oblivious algorithm solving $UTP$ in finite time.  
\end{proof}

\begin{lemma}
\label{lem3:conj}
There exists no algorithm deterministic oblivious algorithm $A$ solving UTP in SYm 
starting from a configuration where the robots form a strict biangular circle.
\end{lemma}
\begin{proof}
Assume by contradiction that there exists a deterministic oblivious algorithm $A$ solving $UTP$ in Sym
starting from a configuration where the robots form a strict biangular circle.  
Assume that the $n$ robots that, initially, the robots in $G_1$ (resp in $G_2$) of the strict biangular 
circle have the same view.  Note that the view of the robots in $G_1$ may be different than the view 
of the robots in $G2$. In such a configuration, the robots are said to be in a \emph{special biangular circle}. 
In the following of the proof, we also assume that if one robot in $G_1$ (resp. $G_2$) becomes active at time $t_j$,
then all the robots in $G_1$ (resp. $G_2$) are active in $t_j$.  

By fairness, at least one robot $r_i$ becomes active at time $t_j$.  Without loss of generality, assume that 
$r_i \in G_1$. By assumption, all the robots in $G_1$ are active in $t_j$.   There are only two cases: 
\begin{enumerate}
  \item The robots $\in G_1$ move such that $G_1(t_{j+1})\cup G_2(t_j)$ do not form a regular $n$-gon.
  Then, assuming that no robot in $G_2$ is active at $t_j$, $G_2(t_j)=G_2(t_{j+1})$.  
  So, $G_1(t_{j+1}) \cup G_2(t_{j+1})$ do not form a regular $n$-gon.  

  \item The  robots $\in G_1$ move such that $G_1(t_{j+1})\cup G_2(t_j)$ form a regular $n$-gon.
  Then, assume that all the robots in $G_2$ are active at $t_j$.  Clearly, the only possibility that 
  at time $t_{j+1}$, the robots form a regular $n$-gon is that $G_2(t_{j+1})$ coincides with $G_2(t_j)$.
  This contradicts Lemma~\ref{lem1:conj} and \ref{lem2:conj}. 
  Thus, $G_1(t_{j+1})\cup G_2(t_{j+1})$ do not form a regular $n$-gon. 
\end{enumerate}

So, in both cases, $G_1(t_{j+1})\cup G_2(t_{j+1})$ do not form a regular $n$-gon.  
Since all the robots in $G_1$ (resp. $G_2$) share the same view and execute the same 
deterministic algorithm $A$, every robot $r_i$ in $G_1$ (resp. $G_2$) moves in the exact 
same way at the same time along the boundary of a same circle. Thus, either $G_1(t_{j+1})\ne G_2(t_{j+1})$ 
or $G_1(t_{j+1})=G_2(t_{j+1})$. In the former case, the robots form a biangular circle at $t_{j+1}$.
The latter case, it would be impossible to separate $G_1$ and $G_2$ in a deterministic manner. 
The lemma is proven by induction. 
\end{proof}

The proof of Conjecture~\ref{conj:DK02} directly follows from Lemma~\ref{lem3:conj}. 
\section{Circle Formation Protocol}
\label{sec:algo}

In this section, we present the main result of this paper.  We first 
provide particular configurations of the system which we use for simplifying 
the design and proofs of the protocol. Next, the protocol is presented. 

\subsection{Definitions and Basics properties}

\begin{definition}[regular $(k,n)$-gon] A cohort of $k$ robots ($0 < k \leq n$) forms 
a regular $(k,n)$-gon if their positions coincide with a regular $n$-gon such that $n-k$ robots 
are missing. 
\end{definition}

\begin{figure*}[!htbp]
\begin{center}
  \begin{minipage}[t]{0.4\linewidth}
    \centering
    \epsfig{file=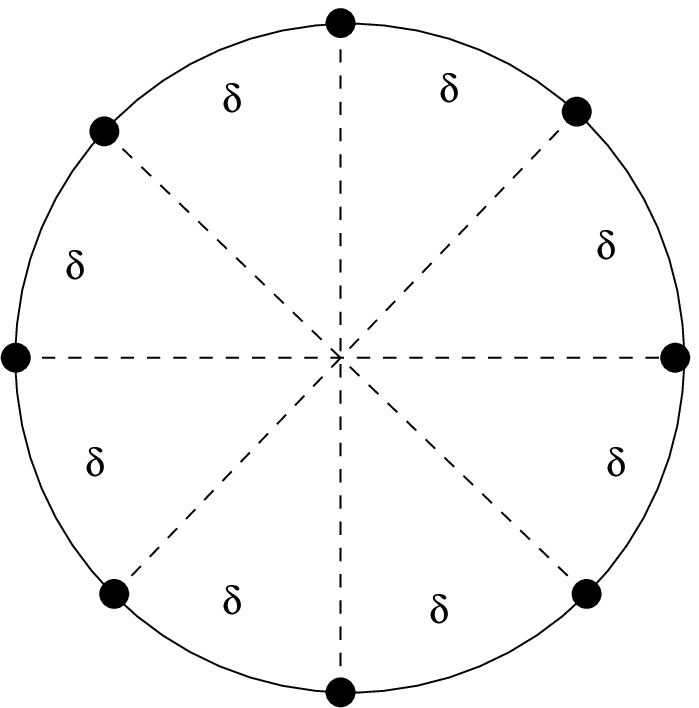, width=0.6\linewidth}\\
    {\footnotesize ($a$) A $8$-gon ($\alpha = \frac{\pi}{4}$)}.
  \end{minipage}%
  \begin{minipage}[t]{0.4\linewidth}
    \centering
    \epsfig{file=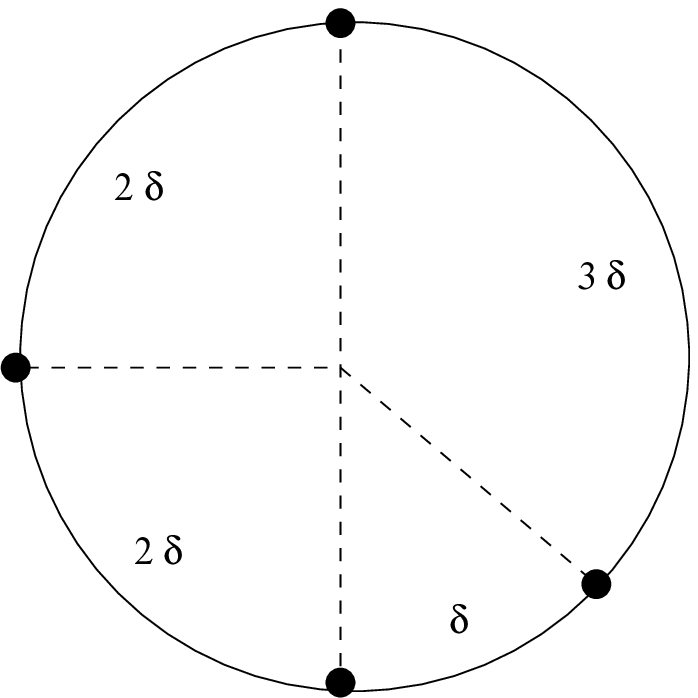, width=0.6\linewidth}\\
    {\footnotesize ($b$) A $(4,8)$-gon obtained removing $4$ robots from the corresponding $8$-gon.} 
  \end{minipage}
\end{center}
 \caption{An example showing a $(k,n)$-gon. \label{fig:ngon}}
\end{figure*}

An example of a $(k,n)$-gon is given in Figure~\ref{fig:ngon}.
Given a $(k,n)$-gon such that $k \geq 2$, if $p$ robots are missing (w.r.t. the corresponding 
$n$-gon) between two adjacent robots, then $\widehat{r O r'}= (p+1) \frac{2\pi}{n}$.
Given a $(1,n)$-gon, then number of missing robot is equal to $n-1$.  Remark that since 
the uniqueness of any circle is guaranteed by passing through $3$ points only, there is an infinity of
circles passing through $1$ or $2$ robots.  So, if $k\leq 2$, then there is an infinity of $(k,n)$-gon
passing through $k$ robots.\\

Let $C_1$ and $C_2$ be two circles having their radius greater than $0$. 
$C_1$ and $C_2$ are said to be {\it concentric} if they share the same center but their radius are different. 
Without lost of generality, in the remainder, given a pair $(C_1, C_2)$ of concentric circles, 
$C_1$ (resp. $C_2$) indicates the 
circle with the greatest radius (resp. smallest radius). 
\begin{definition}[Concentric Configuration]
\label{def:cc}
The system is said to be in a concentric configuration if there exists a pair of concentric circles 
$(C_1,C_2)$ and a partition of the $n$ robots into two subsets $A$ and $B$ such that 
every robot of $A$ (respectively $B$) is located on $C_1$ (resp. $C_2$).
\end{definition}

\begin{figure*}[!htbp]
\begin{center}
  \begin{minipage}[t]{0.4\linewidth}
    \centering
    \epsfig{file=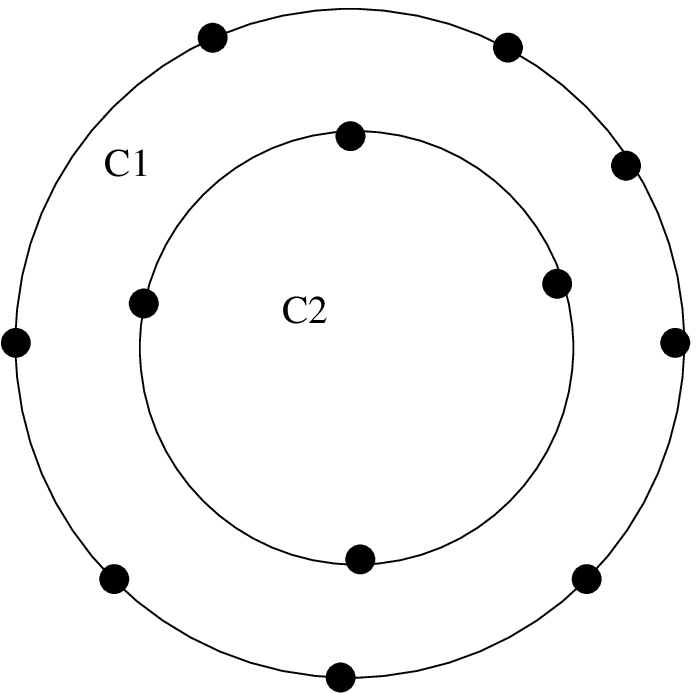, width=0.6\linewidth}\\
    {\footnotesize ($a$) An example of a concentric configuration with $n=12$.} 
  \end{minipage}%
  \begin{minipage}[t]{0.4\linewidth}
    \centering
    \epsfig{file=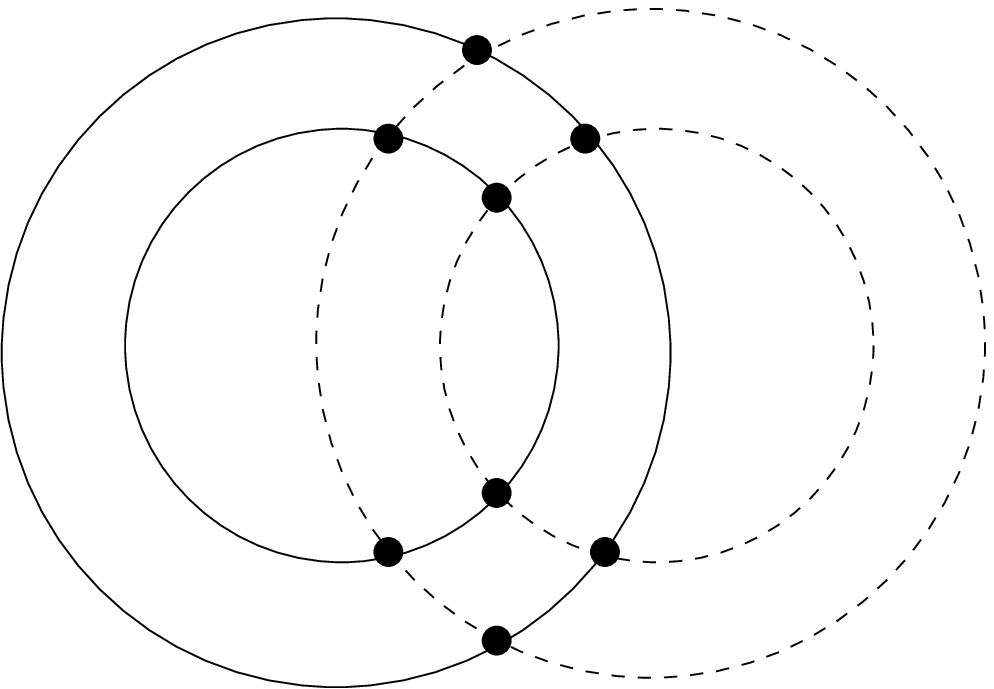, width=0.85\linewidth}\\
    {\footnotesize ($b$) An example showing that the pair of concentric circle may not be unique with $n\leq8$.} 
  \end{minipage}
\end{center}
 \caption{Examples of concentric configurations.  \label{fig:concentric}}
\end{figure*}

\begin{remark}
\label{rem:part}
 $A\ne \emptyset$ and $B\ne \emptyset$.
\end{remark}
\begin{remark}
\label{rem:n9}
If $n \leq 8$, then the pair $(C_1,C_2)$ may not be unique.
\end{remark}
An example illustrated Remark~\ref{rem:n9} is given in Figure~\ref{fig:concentric}.

\begin{lemma}
\label{lem:not}
If the system is in a concentric configuration and if $n > 8$, then there exists a single pair 
$(C_1, C_2)$ in which all the robots are located.
\end{lemma}
\begin{proof}
Assume by contradiction, that the system is in a concentric configuration, $n > 8$ and there exists 
two pairs $\gamma=(C_1, C_2)$ and $\gamma'=(C_1',C_2')$ such that $\gamma \neq \gamma'$ 
(i.e., $C_1 \neq C_1'$, $C_1 \neq C_2'$, $C_2 \neq C_1'$ and $C_2\neq C_2'$) and in which all the robots are located.  
Since two different circles share at most two points,
the pairs $\gamma$ can share at most eight robots with $\gamma'$ (refer to Case~$(b)$ in 
Figure~\ref{fig:concentric}).  Since by assumption $n \geq 9$, there exists at 
least one robot which is located on either $C_1$ or $C_2$, but which is located on neither $C_1'$ nor $C_2'$.
This contradicts the fact that each robot is located either on $C_1'$ or on $C_2'$.
\end{proof}

So, from Lemma~\ref{lem:not}, when the system is in a concentric configuration and $n\geq9$, the pair $(C_1,C_2)$ is unique.  In such a configuration,
given a robot $r$, $proj(r)$ denotes the projection of $r$ on $C_1$, 
i.e., the intersection between the half-line $[c,r)$ and $C_1$, where $c$ is the center of $(C_1,C_2)$. 
Obviously, if $r$ is located on $C_1$, then $proj(r)=r$. We denote by $\Pi$ the projection set of the $n$ robots. 
In a concentric configuration, if $|\Pi|=n$, then the radii passing through the robots on $C1$ split up the disk 
bounded by $C_1$ into {\it sectors}.  


\begin{figure*}[!htbp]
\begin{center}
  \begin{minipage}[t]{0.4\linewidth}
    \centering
    \epsfig{file=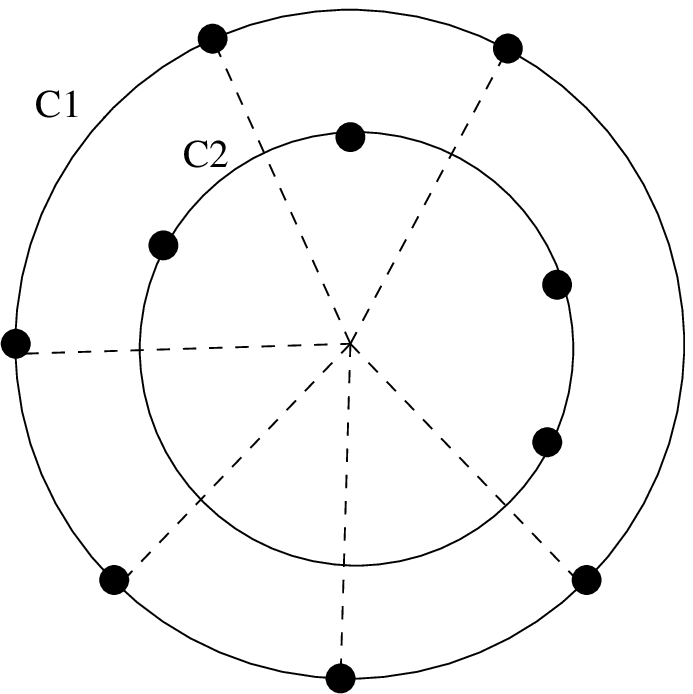, width=0.6\linewidth}\\
    {\footnotesize ($a$)}
  \end{minipage}%
  \begin{minipage}[t]{0.4\linewidth}
    \centering
    \epsfig{file=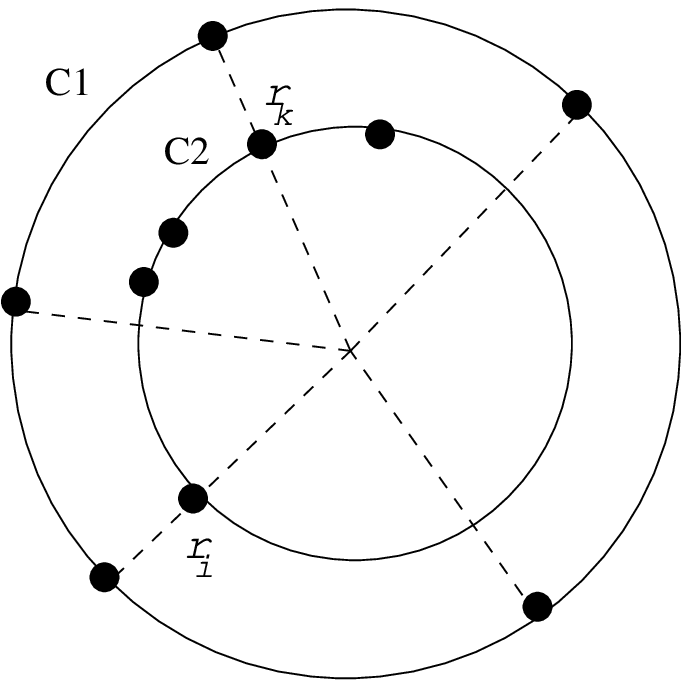, width=0.6\linewidth}\\
    {\footnotesize ($b$)}
  \end{minipage}
\end{center}
\caption{The concentric configuration shown in Case~($a$) is split up into sectors, whereas
  the one in Case~($b$) is not because the some robots on $C1$ are located on the projections of 
  $r_i$ and $r_k$.  \label{fig:sector}}
\end{figure*}


\begin{definition}[quasi $n$-gon] 
A cohort of $n$ robots ($n \geq 9$) forms 
an (arbitrary) quasi n-gon iff the three following conditions hold:
\begin{enumerate}
\item The robots form a concentric configuration divided into sectors; 
\item The robots on $C_1$ form a regular $(k,n)$-gon; 
\item In each sector, if $p$ robots are missing on $C_1$ to form a regular $n$-gon, then 
     $p$ robots are located on $C_2$ in the same sector. 
\end{enumerate}
\end{definition}

\begin{figure*}[!htbp]
\begin{center}
  \begin{minipage}[t]{0.4\linewidth}
    \centering
    \epsfig{file=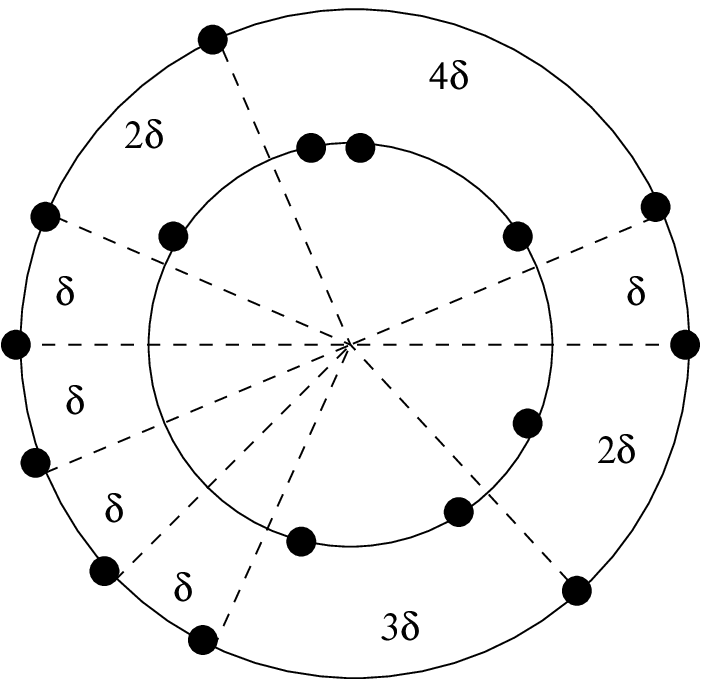, width=0.7\linewidth}\\
    {\footnotesize ($a$) An arbitrary quasi $n$-gon.}
  \end{minipage}%
  \begin{minipage}[t]{0.4\linewidth}
    \centering
    \epsfig{file=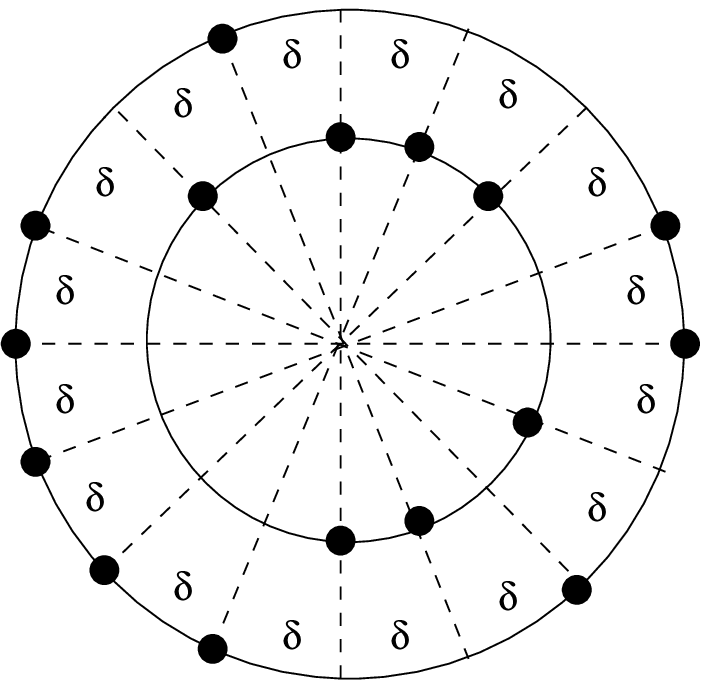, width=0.7\linewidth}\\
    {\footnotesize ($b$) An aligned quasi $n$-gon.}
  \end{minipage}
\end{center}
\caption{Two quasi $n$-gon with $n=16$.\label{fig:QD}}
\end{figure*}

A quasi $n$-gon is said to be \emph{aligned} iff $Pi$ coincide with a regular $n$-gon.
Two quasi $n$-gon are shown in Figure~\ref{fig:QD}, the first one is arbitrary, the other one is
aligned.

\subsection{The Protocol}

Let us consider the overall scheme of our protocol presented in Algorithm~\ref{algo:main}.  
It is mainly based on the particular configurations presented in the previous subsection.  

As mentioned in the introduction, the proposed scheme is combined with the protocol presented
in~\cite{K05} which leads a cohort of $n$ robots from an arbitrary to a 
biangular configuration, with $n\geq 2$. 
In the remainder, we refer to the protocol in~\cite{K05} as 
Procedure~$\AB$---from an {\em A}rbitrary configuration to a {\em B}iangular configuration.  
The model used in~\cite{K05}, called Corda~\cite{P02}, allows more asynchrony 
among the robots than the semi-asynchronous model used in this paper---let us call it $SSM$. 
However, we borrow the following result from~\cite{P02}:
\begin{theorem}\cite{P02}
Any algorithm that correctly solves a problem $P$ in Corda, correctly solves $P$ in $SSM$.
\end{theorem}

The above result means that Procedure~$\AB$ can be used in $SSM$.  
Obviously, Procedure~$\AB$ trivially solves the CFP if the number of robots $n$ is 
odd. 
So, to solve CFP for any number of robots, it
remains to deal with a system in a strict biangular configuration when $n$ is even. \\ 

In the remainder, we consider that the system is in an arbitrary configuration
if the robots do not form either $(1)$ a regular $n$-gon, $(2)$ a quasi $n$-gon, 
or $(3)$ a strict biangular circle.  
Let us describe the general scheme provided by Algorithm~\ref{algo:main}.  

Procedure~$\AB$ excluded, the protocol mainly consists of three procedures.  
The first one, called Procedure~$\aQN$ is used when the system form an aligned quasi $n$-gon.  It leads 
the system into a regular $n$-gon.  The aim of Procedure~$\QaQ$ is to transform
the cohort from an arbitrary quasi $n$-gon into an aligned quasi $n$-gon.  The last procedure, 
Procedure~$\BQ$, is used when the robots form a biangular circle and arranges them
into either a regular $n$-gon or an arbitrary quasi $n$-gon, 
depending on the synchrony of the robots.
The details of those procedures are given in the remainder of this section.

Let us explain how the procedures are used by giving the overall scheme of 
Algorithm~\ref{algo:main}.  Starting from an arbitrary configuration, using Procedure~$\AB$, 
the system is eventually in a biangular circle. If $n$ is odd, then the robots form a regular $n$-gon 
and the system is done.  Otherwise ($n$ is even), the robots form either a regular $n$-gon
or a strict biangular circle. Starting from the latter case, each robot executes Procedure~$\BQ$.  
As mentioned above, the resulting configuration can be either a regular $n$-gon or a quasi $n$-gon.
>From a quasi $n$-gon, the robots execute either Procedure~$\aQN$ or Procedure~$\QaQ$, 
depending on whether the quasi $n$-gon is aligned or not.

Both procedures~$\aQN$ and~$\QaQ$ require no ambiguity on the concentric configuration 
forming the quasi $n$-gon, i.e $n\geq9$. 
However, since $\aQN$ and~$\QaQ$ are called when $n$ is even only,
only the cases $n=4,6$ and $8$ are not solved by our algorithm. 
So, in the remainder, we assume that $n \notin \{4,6,8\}$.
Finally, starting from an aligned quasi $n$-gon, the resulting configuration of 
the execution of Procedure~$\aQN$ is a regular $n$-gon.
Otherwise, the quasi $n$-gon becomes aligned by executing Procedure~$\QaQ$.

\begin{algo}[htb]
\begin{small}
\begin{tabbing}
  xxxxx \= xxxxx \= xxxxx \= xxxxx \= xxxxx \= xxxxx \= xxxxx \= xxxxx \= xxxxx \= \kill 
n:= the number of robots;\\
\IF{$n$ is even}\\ 
\THEN \>\IF{the robots do not form a regular $n$-gon}\\
      \> \THEN \> \IF{the robots form a quasi $n$-gon}\\ 
                  \> \THEN \> \IF{the robots form an aligned quasi $n$-gon}\\
                  \>       \> \THEN Execute $\aQN$;\\ 
                  \>       \> \ELSE Execute $\QaQ$;\\
                  \> \ELSE \> \IF{the robots form a strict biangular circle}\\
                  \>       \> \THEN Execute $\BQ$;\\
                  \>       \> \ELSE Execute $\AB$; \\
\ELSE \> Execute $\AB$;
\end{tabbing}
\end{small}
\caption{Procedure $\Ngon$ for any $r_i$ in a cohort of $n$ robots ($n\neq 4$, $6$, or $8$). \label{algo:main}}
\end{algo}
\begin{theorem}
\label{th:ngon}
Procedure~$\Ngon$ is a deterministic Circle Formation Protocol for any number $n$ of robots such that
$n \notin \{4,6,8\}$.  
\end{theorem}

The above theorem follows from Procedure~$\Ngon$ (Algorithm~\ref{algo:main}, \cite{K05}, 
Lemmas~\ref{lem:aQN}, \ref{lem:QaQ}, and~\ref{lem:BQ}.
In the remainder of this section, the procedures and the proofs of the three above lemmas are presented 
in separate paragraphs. 

\paragraph{Procedure~$\aQN$.}
Starting from an aligned quasi $n$-gon, each robots on $C_2$ needs to move toward its projection on $C_1$
whereas it is required that any robot on $C_1$ remains at the same position because it is located on 
its projection.  This obvious behavior is made of the following single instruction:
%
$$
  \mbox{move to }proj(r_i)
$$
Since we have $n\geq9$ in quasi $n$-gon, from Lemma~\ref{lem:not}, the pair 
$(C_1,C_2)$ is unique.  Moreover, it remains unchanged while the regular $n$-gon is not formed.
So, the following result holds:
\begin{lemma}
\label{lem:aQN}
Starting from an aligned quasi $n$-gon, Procedure~$\aQN$ solves the Circle Formation Problem.
\end{lemma}

\paragraph{Procedure~$\QaQ$.}

The idea behind Procedure~$\QaQ$ consists in changing a quasi $n$-gon into an aligned quasi $n$-gon
by arranging the robots on $C_2$ in each sector---refer to Figure~\ref{fig:QD}. 

In the following of the paragraph, denote a quasi $n$-gon by the corresponding pair of 
concentric circles $(C_1,C_2)$.  Two  quasi $n$-gons $(C^\alpha_1,C^\alpha_2)$ and 
$(C^\beta_1,C^\beta_2)$ are said to be \emph{equivalent} if
$C^\alpha_1 = C^\beta_1$, $C^\alpha_2 = C^\beta_2$ and the positions of the robots on 
$C^\alpha_1$ and $C^\beta_1$ are the same ones. In other words, the only allowed 
possible difference between two equivalent 
 quasi $n$-gons $(C^\alpha_1,C^\alpha_2)$ and $(C^\beta_1,C^\beta_2)$ 
is different positions of robots between $C^\alpha_2$ and $C^\beta_2$ in each sector.

Procedure~$\QaQ$ is shown Algorithm~\ref{algo:QaQ}.  This procedure assumes that
the initial configuration is an arbitrary quasi $n$-gon.  In such a configuration, 
we build, a partial order among the robots on $C_2$ belonging to a common sector 
to eventually form an aligned quasi $n$-gon.

\begin{algo}[htb]
\begin{small}
\begin{tabbing}
  xxxxx \= xxxxx \= xxxxx \= xxxxx \= xxxxx \= xxxxx \= xxxxx \= xxxxx \= xxxxx \= \kill 
$C_1:= \mbox{greatest concentric circle}$; $C_2:= \mbox{smallest concentric circle}$; \\
\IF {$r_i$ are located on $C_2$} \\
\THEN \> $MySector:= \mbox{sector wherein }r_i \mbox{ is located}$;\\ 
      \> $PS:= FindFinalPos(Mysector)$;\\
      \> $FRS:= \mbox{set of robots in MySector which are not located on a position in } PS$;\\
      \> \IF{$FRS \ne \emptyset$}\\
      \> \THEN \> $EFR:= ElectFreeRobots(FRS)$;\\
      \>       \> \IF{$r_i \in EFR$} 
            \THEN move to Position $Associate(r_i)$; 
\end{tabbing}
\end{small}
\caption{Procedure~$\QaQ$ for any robot $r_i$ in an arbitrary quasi $n$-gon \label{algo:QaQ}}
\end{algo}

Let $p_1,\ldots,p_s$ be the final positions on $C_2$ in the sector $S$ in 
order to form the aligned quasi $n$-gon. Let $B_1,B_2$ the two points located 
on $C_2$ at the boundaries of $S$. Of course, if only one robot is located on $C_1$ 
(i.e. there exists only one sector), then $B_1=B_2$.   
For each $i \in 1\ldots s$, $p_i$ is the point on $C_2$ in $S$ such that 
$\widehat{B_1 O p_i} = \frac{2k\pi}{n}$, $p_i\ne B_1$ and $p_i\ne B_2$. 
Clearly, while the distributed system remains in an equivalent
 quasi $n$-gon, all the final 
positions remain unchanged for every robot. 
A final position $p_i$,  $i \in 1\ldots s$, is said to be \emph{free} if no robot takes 
place at $p_i$.  Similarly, a robot $r_i$ on $C_2$ in $S$ is called a \emph{free} robot if its current position 
does not belong to $\{p_1, \ldots, p_s\}$.

Define Function $FindFinalPos(S)$ which returns the set of final positions on $C_2$ in $S$ 
with respect to $B_1$.  Clearly, in $S$ all the robots compute the same set of final positions, stored in $PS$.  
Each robot also temporarily stores the set of free robots in the variable called $FRS$.  
Of course, since the robots are oblivious, each active robot on $C_2$ re-compute $PS$ and $FRS$ each time 
Procedure~$\QaQ$ is executed.
Basically, if $FRS = \emptyset$ all the robots occupy a final position in the sector $S$.  
Otherwise, the robots move in waves to the final positions in their sector following the order 
defined by Function $ElectFreeRobots()$.  In each sector, the elected 
robots are the closest free robots from $B_1$ and $B_2$.  
Clearly, the result of Function $ElectFreeRobots()$ return the same set of robots for 
every robot in the same sector.  Also, the number of elected robots is at most
equal to $2$, one for each point $B_1$ and $B_2$.  Note that it can be 
equal to $1$ when there is only one free robot, i.e., when only one robot in $S$ did
not reach the last free position.

Function $Associate(r)$ assigns a unique free position to an elected robot as follows:\newline
\noindent If $ElectFreeRobots()$ returns only one robot $r_i$, then $r_i$ is associated to the 
single free remaining position $p_i$ in its sector.  This allows $r_i$ to move to $p_i$.  
If $ElectFreeRobots()$ returns a pair of robots $\{r_i,r_{i'}\}$ ($r_i \ne r_{i'}$), 
then the closest robot to $B_1$ (respectively, $B_2$) is associated with the closest position to 
$B_1$ (resp., $B_2$) in $S$.  
Note that, even if the robots may have opposite clockwise directions, $r_i$, $r_{i'}$, and their associated 
positions are the same for every robot in $S$. 

\begin{lemma}
\label{lem:oc}
According to Procedure~$\QaQ$, if the robots are in a quasi $n$-gon at time $t_j$ ($j\geq0$), then
at time $t_{j+1}$, the robots are in an equivalent quasi $n$-gon.
\end{lemma}
\begin{proof}
By assumption, at each time instant $t_j$,
at least one robot is active.  So, by fairness, starting from a quasi $n$-gon, at least one robot executes 
Procedure~$\QaQ$.  Assume first that no robot executing Procedure~$\QaQ$ moves from $t_j$ to $t_{j+1}$.  
In that case, since the robots are located on the same positions at $t_j$ and at $t_{j+1}$, the robots 
are in the same quasi $n$-gon at $t_{j+1}$. Hence, the robots remains in an equivalent quasi $n$-gon 
seeing that any quasi $n$-gon is  equivalent to itself. So, at least one robot moves from $t_j$ to $t_{j+1}$.
However, in each sector at most two robots are allowed to move toward distinct free positions on $C_2$ only 
inside their sector.  Thus, the robots remains in an equivalent quasi $n$-gon.
\end{proof}


The following lemma follows from Lemma~\ref{lem:oc} and fairness:
\begin{lemma}
\label{lem:QaQ}
Procedure~$\QaQ$ is a deterministic algorithm transforming an arbitrary quasi $n$-gon into an aligned $n$-gon 
in finite time.
\end{lemma}

\begin{figure*}[!htbp]
\begin{center}
  \begin{minipage}[t]{0.38\linewidth}
    \centering
    \epsfig{file=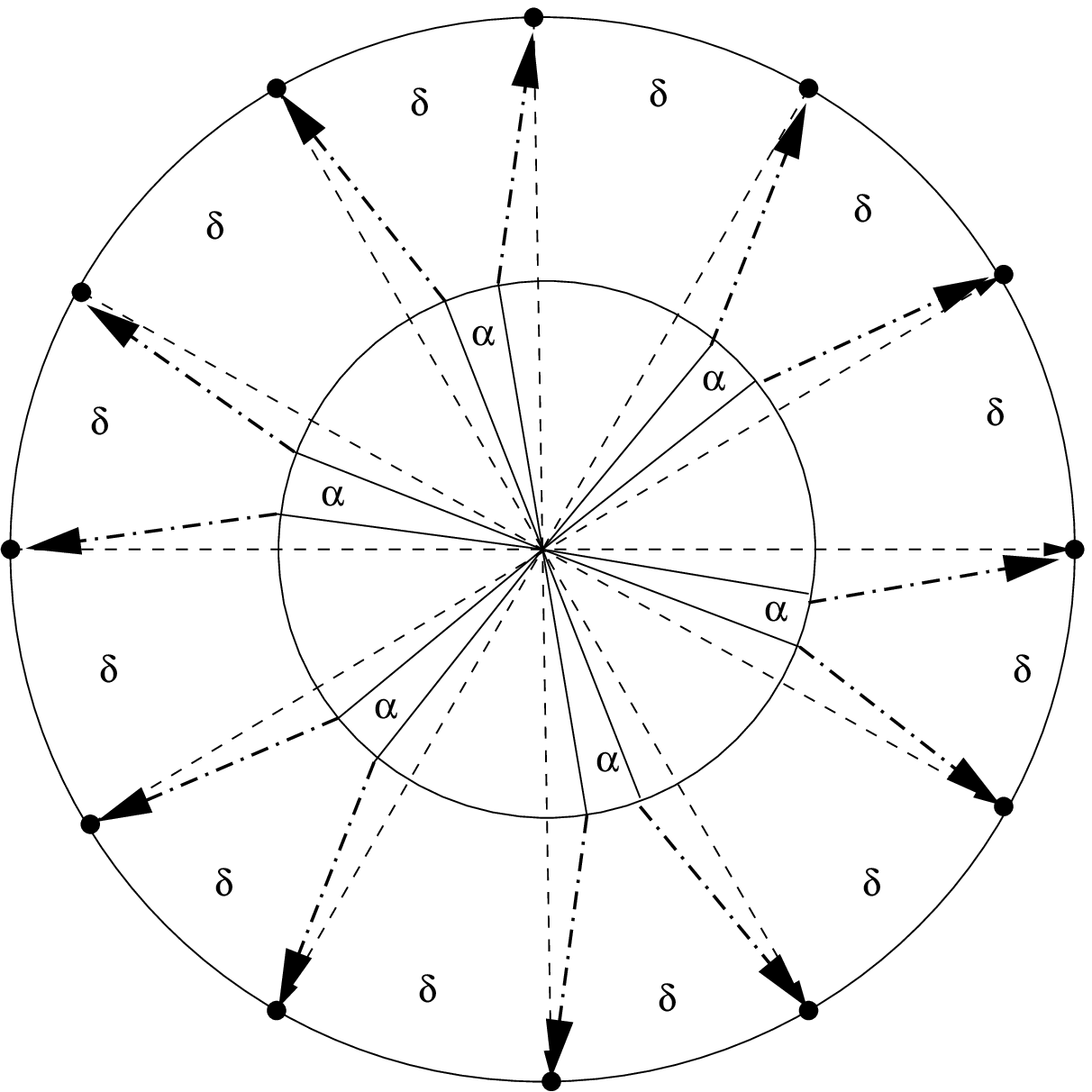, width=0.9\linewidth}\\
    {\footnotesize ($a$) If all the robots are active at $t_j$, then the robots form a regular $n$-gon at $t_{j+1}$.}
  \end{minipage}%
  \begin{minipage}[t]{0.38\linewidth}
    \centering
    \epsfig{file=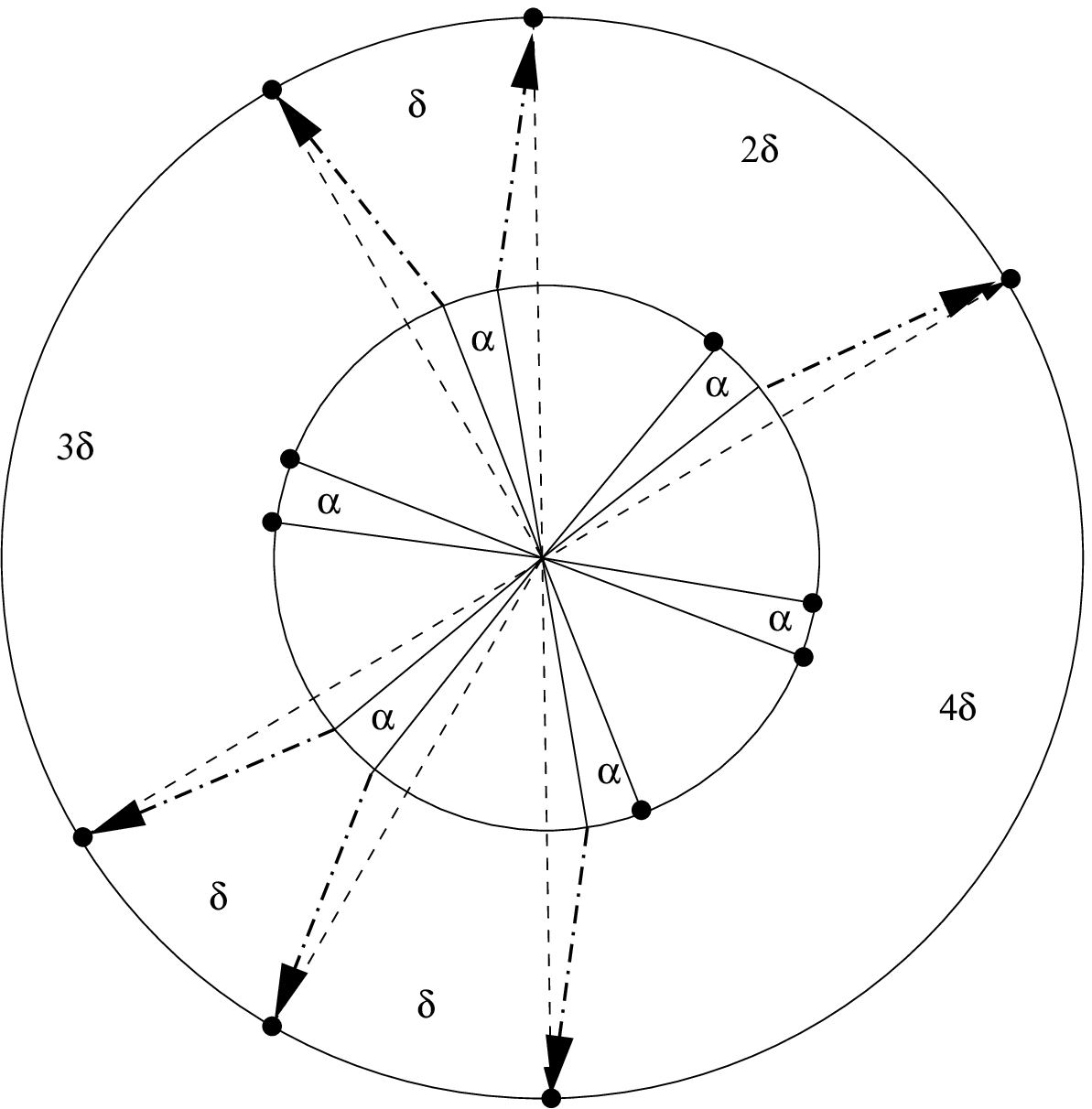, width=0.9\linewidth}\\
    {\footnotesize ($b$) If some robots are inactive at $t_j$, then the robots form a quasi $n$-gon at $t_{j+1}$.}
  \end{minipage}
\end{center}
\caption{An example showing the principle of Procedure~$\BQ$.\label{fig:explain}}
\end{figure*}

\paragraph{Procedure $\BQ$.} 
We assume that initially, the robots from a strict biangular circle.  
In such a configuration, every active robots $r_i$ apply the following scheme:  

\begin{enumerate}
\item 
  Robot $r_i$ computes the concentric circle $C'$ whose the radius is twice 
  the radius of the strict biangular circle $C$;
\item
  Robot $r_i$ considers its neighbor $r_{i'}$ such that $\widehat{r_i O r_{i'}}=\alpha$
  and $r_i$ moves away from $r_i'$ to the position $p_i(t_{j+1})$ on $C'$ 
  with an angle equal to $\frac{\pi}{n}-\frac{\alpha}{2}$. More precisely, 
  $\widehat{p_i(t_{j+1}) O p_{i}(t_j)} = \frac{\pi}{n} - \frac{\alpha}{2}$ and 
  $\widehat{p_i(t_{j+1}) O p_{i'}(t_j)} = \frac{\pi}{n} + \frac{\alpha}{2}$
  ---refer to Figure~\ref{fig:explain}.
\end{enumerate}

Let us consider two possible behaviors depending on the synchrony of the robots.  
\begin{enumerate}
\item 
  Assume that every robot in the strict biangular circle is active at time $t_j$.  
  In that case, at $t_{j+1}$, the robots form a regular $n$-gon---see Case $(a)$ in 
  Figure~\ref{fig:explain}.  Indeed, there are two cases: 
  \begin{enumerate}
  \item
    If $\widehat{p_i(t_j) O p_{i'}(t_j)}=\alpha$, then 
	  $\widehat{p_i(t_{j+1}) O p_{i'}(t_{j+1})} = \alpha + 2(\frac{\pi}{n}-\frac{\alpha}{2})$.\newline
    So, in that case, $\widehat{p_i(t_{j+1}) O p_{i'}(t_{j+1})} = 2 \frac{\pi}{n}$.  
  \item
    If $\widehat{p_i(t) O p_{i'}(t_j)} = \beta$, then 
          $\widehat{p_i(t_{j+1}) O p_{i'}(t_{j+1})} = \beta - 2(\frac{\pi}{n}-\frac{\alpha}{2})$.\newline
    So, in that case, $\widehat{p_i(t_{j+1}) O p_{i'}(t_{j+1})} = \beta - 2\frac{\pi}{n} + \alpha$, 
    which also equal to $\beta - 4\frac{\pi}{n}+\alpha+2\frac{\pi}{n}$. 
    From Remark~\ref{rem:biangular}, we know that $\beta = 4\frac{\pi}{n}-\alpha$. 
    Hence, $\widehat{p_i(t_{j+1}) O p_{i'}(t_{j+1})} = \beta -\beta +2\frac{\pi}{n}$,
    which is equal to $2\frac{\pi}{n}$.
  \end{enumerate}
  Note $(1)$ the trajectories of the robots do not cross between them, and 
  $(2)$ all the angles $\alpha$ (resp. $\beta$) increases up (resp. decrease down) to $\frac{2\pi}{n}$.
\item 
  Assume that some robots, in the strict biangular circle, are not active at time $t_j$.
  In that case, only a subset of robots move toward $C'$ from $t_j$ to $t_{j+1}$. 
  Then, the robots form a quasi $n$-gon at time $t_{j+1}$---see Case $(b)$ in 
  Figure~\ref{fig:explain}. 
  Indeed at $t_{j+1}$, the robots are in a concentric configuration where $C_1$ is $C'$ and $C_2$ is 
  the initial circle $C$ (i.e the biangular circle at time $t_j$). 
  Furthermore on $C_1$, the robots form a regular $(k,n)$-gon where $n-k$ represent the subset of robots 
  which remain inactive at time $t_j$.  
\end{enumerate}

To show that, if the system eventually do not form a regular $n$-gon, we need to prove that it eventually
form a quasi $n$-gon.  Following the above explanations, in remains to show that, in the above second case, 
the configuration is sliced into sectors at time $t_{j+1}$ such that, in each sector, the missing robots 
on $C_1$ are located on $C_2$.

\begin{lemma}
\label{lem:BQ1}
Using Procedure~$\BQ$, if all the robots are in strict biangular circle at time $t_j$, then 
the configuration is sliced into sectors at $t_{j+1}$ when the $n$-gon is not formed.
\end{lemma}
  
\begin{proof}
As already stated previously, the robots are in concentric configuration at time $t_{j+1}$. 
Moreover, at $t_j$, the robots are in a strict biangular circle such that $\alpha+\beta=\frac{4\pi}{n}$. 
Since the biangular circle is strict, without loss of generality, we can assume that 
$\alpha<\beta$ with $0<\alpha<\frac{2\pi}{n}$ and $\frac{2\pi}{n}<\beta<\frac{4\pi}{n}$. 
 
Assume, by contradiction, that there exists one robot $r_i$ on $C_2$ located on the radius 
passing through any robot $r_{i'}$ on $C_1$ at $t_{j+1}$. This implies that at $t_j$, 
$\widehat{r_i O r_{i'}}= \frac{\pi}{n}-\frac{\alpha}{2}$, i.e., 
the angle whose $r_{i'}$ moved away from $r_{i}$ on $C'$ from $t_j$ to $t_{j+1}$. 
Furthermore, at $t_j$, $r_{i'}$ is active and $r_i$ is inactive.  
Note that $\widehat{p_i(t_j) O p_{i'}(t_j)}$ is either equal to $\alpha$ or $\beta$.  Thus, 
either $\frac{\pi}{n}-\frac{\alpha}{2}=\alpha$ or $\frac{\pi}{n}-\frac{\alpha}{2}=\beta$. 
However, $\frac{\pi}{n}-\frac{\alpha}{2}<\frac{2\pi}{n}$, and $\frac{2\pi}{n}<\beta<\frac{4\pi}{n}$. 
Hence, $\frac{\pi}{n}-\frac{\alpha}{2}=\alpha$, and then $\widehat{p_i(t_j) O p_{i'}(t_j)} = \alpha$. 
By executing Procedure~$\BQ$, $r_{i'}$ moves away from $r_i$ with an angle $\frac{\pi}{n}-\frac{\alpha}{2}$,
where $0<\frac{\pi}{n}-\frac{\alpha}{2}<\frac{2\pi}{n}$. 
Since $r_i$ is inactive we have $\widehat{p_i(t_{j+1}) O p_{i'}(t_{j+1})}= (\frac{\pi}{n}-\frac{\alpha}{2})+\alpha$.
Furthermore, Procedure~$\BQ$ is called only when $n\geq9$, and thus, 
we have  $0<(\frac{\pi}{n}-\frac{\alpha}{2})+\alpha<\frac{2\pi}{9}+\frac{2\pi}{9}=\frac{4\pi}{9}$ and 
$0<\widehat{p_i(t_{j+1}) O p_{i'}(t_{j+1})}< \frac{4\pi}{9}$. 
Thus, at $t_{j+1}$, $r_i$ and $r_{i'}$ are not on the same radius. A contradiction.
\end{proof}

\begin{lemma}
\label{lem:BQ2}
Using Procedure~$\BQ$, if all the robots form a strict biangular circle at time $t_j$, then 
in each sector, the missing robots on $C_1$ are located on $C_2$ at $t_{j+1}$ when the $n$-gon is not formed.
\end{lemma}

\begin{proof}
Clearly, when all the robots are active and move simultaneously by applying our method, 
the trajectories do not cross between them (see Figure \ref{fig:explain}). 
Assume by contradiction, that at time $t_{j+1}$, there exists any sector with 
one extra robot $r$. If all the robots have been active at time $t_j$, $r$ would 
have crossed any other trajectory in order to form a regular $n$-gon.  A contradiction.  
\end{proof}

The following lemma directly follows from the algorithm, Lemmas~\ref{lem:BQ1} and~\ref{lem:BQ2}:
\begin{lemma}
\label{lem:BQ}
Procedure~$\BQ$ is a deterministic algorithm transforming a biangular circle into either a regular
$n$-gon or quasi $n$-gon in finite time.
\end{lemma}

\section{Concluding Remarks}
\label{sec:conclu}

In this paper, we studied the problem of forming a regular $n$-gon with a cohort of $n$ robots (CFP). 
We first shown that it is impossible to obtain a regular $n$-gon in a deterministic way only by 
moving the robots along the circle on which all of them take place.  
Next, we presented a new approach for this problem based on concentric circles formed by the 
robots.  Combined with the solution in~\cite{K05}, our solution works with any number of robots $n$
except if $n=4$,$6$ or $8$.  The main reasons that $n$ must be different from $4$, $6$ or $8$ comes 
from the fact that the robots may confuse in the 
recognition of the particular configurations if $n$ is lower than $9$.  The CFP remains open for 
these three special cases.  In a future work, we would like to investigate CFP in 
a weakest model such that Corda. 
   
\bibliographystyle{alpha}
\bibliography{ngon}
\end{document}